\documentclass[sigconf,nonacm,natbib=false]{acmart}
\newcommand\vldbdoi{XX.XX/XXX.XX}

\newcommand\vldbavailabilityurl{http://vldb.org/pvldb/format_vol14.html}

\usepackage{verbatim}
\usepackage{multirow}
\usepackage{textcomp}
\usepackage{booktabs}
\usepackage{amsmath}
\usepackage{bbm}
\usepackage{enumitem}
\usepackage{setspace}
\usepackage[capitalise]{cleveref}
\usepackage{subcaption}
\usepackage[%
backend=bibtex,%
bibstyle=ACM-Reference-Format,%
citestyle=numeric-comp,%
natbib=true,sorting=none]{biblatex}
\usepackage{appendix}

\usepackage{algorithm}
\usepackage[noend]{algpseudocode}
\makeatletter
\renewcommand{\ALG@beginalgorithmic}{\small}
\makeatother
\algnewcommand\Input{\textbf{Input:\space}}
\algnewcommand\Output{\textbf{Ouput:\space}}
\algnewcommand\Let{\textbf{let\space}}

\algnewcommand\algorithmicforeach{\textbf{for each}}
\algdef{S}[FOR]{ForEach}[1]{\algorithmicforeach\ #1\ \algorithmicdo}

\usepackage{tikz}
\usepackage{pgf}

\newcommand{\COMMENT}[1]{}

\usepackage{glossaries}

\newacronym{hybridtte}
{HybridTTE/TSE}
{Hybrid Travel Time and Travel Speed Estimation }

\newacronym{hdse}
{HDSE}
{Hybrid Driving Speed Estimation}

\newacronym{aha}
{UniTE}
{a Unifying approach to Travel time and speed Estimation}

\newacronym{agha}
{G-UniTE}
{Gaussian UniTE}

\newacronym{gru}
{GRU}
{Gated Recurrent Unit}

\newacronym{nll}
{NLL}
{Negative Log-Likelihood}

\newacronym{gps}
{GPS}
{Global Positioning System}

\newacronym{osm}
{OSM}
{OpenStreetMap}

\newacronym{rne}
{RNE}
{Road Network Embedding}

\newacronym{dfs}
{DFS}
{Depth-First Search}

\newacronym{elu}
{ELU}
{Exponential Linear Unit}

\newacronym{relu}
{ReLU}
{Rectified Linear Unit}

\newacronym{bfs}
{BFS}
{Breadth-First Search}

\newacronym{gcn}
{GCN}
{Graph Convolutional Network}

\newacronym{rfn}
{RFN}
{Relational Fusion Network}

\newacronym{mlp}
{MLP}
{Multi-Layer Perceptron}

\newacronym{tsne}
{t-SNE}
{t-distributed Stochastic Neighbor Embedding}

\newacronym{mape}
{MAPE}
{Mean Absolute Percentage Error}

\newacronym{mae}
{MAE}
{Mean Absolute Error}

\newacronym{mse}
{MSE}
{Mean Squared Error}

\newacronym{mpe}
{MPE}
{Mean Percentage Error}

\usepackage[]{todonotes}
\presetkeys{todonotes}{inline}{}

\newcommand{\christian}[1]{
  \todo[inline, color=blue!30!white]{\textbf{Christian}: #1}
}

\presetkeys{todonotes}{disable}{}
\creflabelformat{equation}{#2\textup{#1}#3}

\newcommand{\interval}[2]{\text{[#1; #2)}}
\usepackage{layouts}
\usepackage{svg}

\COMMENT{
  \begin{abstract}
Travel time or travel speed estimation is central to many intelligent transportation applications.
Existing approaches to travel time and travel speed estimation are either function-fitting or aggregation-based and represent a generalizability-accuracy trade-off.

Function-fitting approaches learn a mapping function from feature vectors of, e.g., routes, to estimates which allows them to generalize to unseen routes.
However, the mapping function is imperfect in practice, resulting in poor accuracy.
Aggregation-based approaches instead aggregate historical data to make estimates over, e.g., routes, which allows them to achieve a very high accuracy given sufficient availability of such data.
However, they rely on inaccurate heuristics when there is not sufficient data available, resulting in poor generalizability.

  In this work, we present \gls{aha}, a framework that combines function-fitting and aggregation-based approaches in a unified approach to achieve both the generalizability of function-fitting approaches and the accuracy of aggregation-based approaches.
  We evaluate \gls{aha} empirically and find that a unified approach achieves $12.93 \text{-} 32.18\%$ and $8.34 \text{-} 18.55\%$ better performance in terms of travel speed distribution modeling and travel time point estimation, respectively, over using a function-fitting or aggregation-based approach alone.
\end{abstract}
}
\begin{abstract}
Travel time or speed estimation are part of many intelligent transportation applications. Existing estimation approaches rely on either function fitting or aggregation and represent different trade-offs between generalizability and accuracy.

Function-fitting approaches learn functions that map feature vectors of, e.g., routes, to travel time or speed estimates, which enables generalization to unseen routes. However, mapping functions are imperfect and offer poor accuracy in practice. Aggregation-based approaches instead form estimates by aggregating historical data, e.g., traversal data for routes. This enables very high accuracy given sufficient data. However, they rely on simplistic heuristics when insufficient data is available, yielding poor generalizability.

  We present \gls{aha} that combines function-fitting and aggregation-based approaches into a unified framework that aims to achieve the generalizability of function-fitting approaches and the accuracy of aggregation-based approaches. An empirical study finds that an instance of UniTE can improve the accuracies of travel speed distribution and travel time estimation by {$40$--$64\%$} and {$3$--$23\%$}, respectively, compared to using function fitting or aggregation alone.
\end{abstract}

\begin{document}
\title{UniTE - The Best of Both Worlds:\\ Unifying Function-Fitting and Aggregation-Based Approaches to Travel Time and Travel Speed Estimation}

\author{Tobias Skovgaard Jepsen}
\orcid{0000-0002-9233-781X}
\affiliation{%
  \institution{Aalborg University}
  \department{Department of Computer Science}
}
\email{tsj@cs.aau.dk}

\author{Christian S. Jensen
\orcid{0000-0002-9697-7670}}
\affiliation{%
  \institution{Aalborg University}
  \department{Department of Computer Science}
}
\email{csj@cs.aau.dk}

\author{Thomas Dyhre Nielsen}
\orcid{0000-0002-4823-6341}
\affiliation{%
  \institution{Aalborg University}
  \department{Department of Computer Science}
}
\email{tdn@cs.aau.dk}

\maketitle
\title{UniTE - The Best of Both Worlds: Unifying Function-Fitting and Aggregation-Based Approaches to Travel Time and Travel Speed Estimation}

\section{Introduction}\label{sec:introduction-paper5}
Estimation of travel time or speed is central to many intelligent transportation applications~\citep{realtime-speed-pred} such as trajectory analysis~\citep{stuttgart}, annotating road segments with travel times~\citep{yang2013using,zheng2013time}, and traffic forecasting~\citep{yu2017spatio}.
This often concerns routes in a road network, with road segments being special cases.
For clarity of presentation, we thus assume that estimation is done for routes in the remainder of the paper, but our work is also relevant to other kinds of data, e.g., location-based travel speed forecasting using loop detectors~\citep{cui2019traffic}.
Existing approaches to travel time and speed estimation can be categorised as either function-fitting%
~\citep{wei2020spatial,lu2020st,ge2020global,zhang2020novel,zhang2020graph,zhang2020network, yin2020attention,lee2020predicting,guo2019attention,cui2019traffic,yu2017spatio,htte,stad,zheng2013time,fu2020estimation,barnes2020bustr,lan2019travel,wu2019deepeta,shen2019tcl,hu2019stochastic,hu2020stochastic,taoyang2019deepist,xi2019path,compacteta,rade2018wedge,zheng2018learning,wang2018will,wang2014travel,yang2013using,workshop}
 or aggregation-based%
~\citep{pace,hu2017enabling,dai2016path,yuan2011tdrive} approaches.

Function-fitting approaches fit a function $f$ with parameters $\psi$ that maps feature vector representations of routes, to travel time and speed estimates.
The parameters $\psi$ are found by using input-output pairs from historical data and minimizing the discrepancy between the mapping's output and the expected output.

Aggregation-based approaches use historical travel time or speed data to compute corresponding at estimates of the travel speed or time for routes, often for different time-of-day intervals.
Estimates are typically given as histograms~\citep{pace}, e.g., as parameters that denote heights of bins in equi-width histograms.

\begin{figure}[t]
  \begin{tikzpicture}
    \hspace*{-0.05cm}
\node[inner sep=0pt] (russell) at (0,0)
    {\includegraphics[height=4.25cm,trim=13cm 0cm 6cm 0cm, clip]{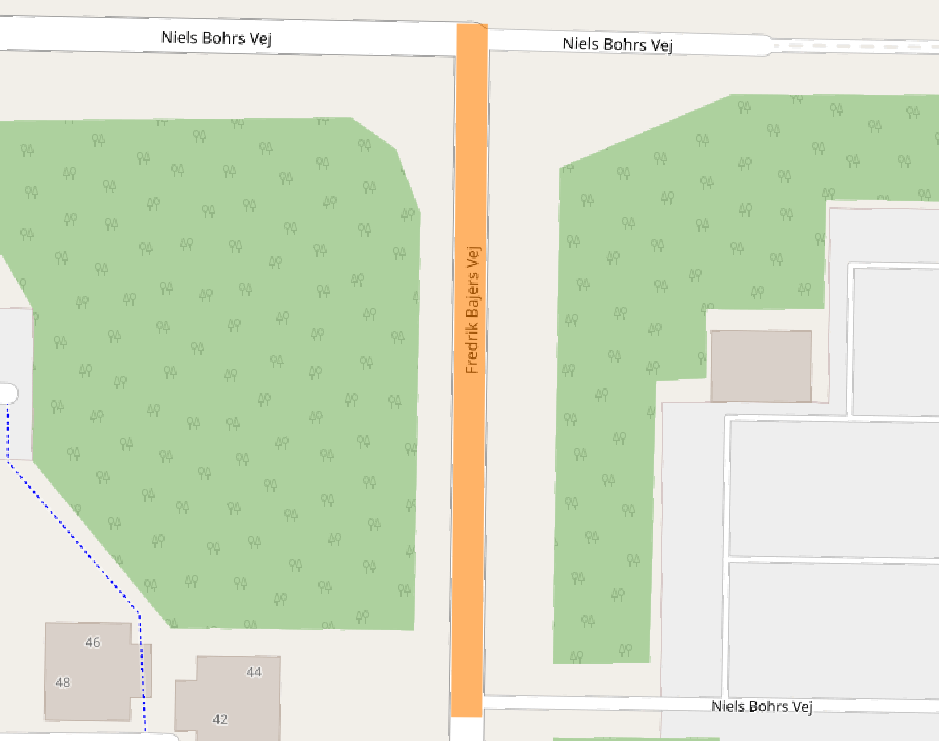}};
\node[inner sep=0pt] (whitehead) at (4,0)
    {\includegraphics[height=4.25cm, trim=0cm 0.42cm 0cm 0.42cm, clip]{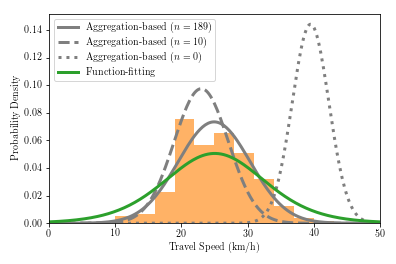}};
\end{tikzpicture}
  \caption{
    A road segment (left) and its ground-truth and estimated travel-speed distributions (right).
    \label{fig:compelling-example-paper5}}
\end{figure}

Function-fitting and aggregation-based approaches represent different trade-offs between estimation generalizability and accuracy.
This trade-off is illustrated in \cref{fig:compelling-example-paper5}.
As shown in the figure, aggregation-based approaches can provide the most accurate estimates given a sufficiently representative sample of size $n=189$.
However, when $n=10$ it underestimates the mean and variance since the data sample is not representative and when $n=0$ it relies on a heuristic.
The heuristic overestimates the mean considerably and also underestimates the variance.
Unlike aggregation-based approaches, function-fitting approaches are unaffected by the representativeness and availability of the data at estimation time.
Instead, it extrapolates the travel speed distribution from other road segments with similar feature representations.
Thus, function-fitting approaches are highly reliant on the availability of high quality feature representations which are typically not easily accessible~\citep{workshop,rfnlong}.
In this case, it estimates the mean well, but overestimates the variance.

As illustrated in \cref{fig:compelling-example-paper5}, function-fitting and aggregation-based approaches perform well in different situations.
However, choosing which approach is preferable in a particular situation is difficult in practice since the representability of the samples and the performance of the function-fitting approach are not known a priori.
The sample representability is particularly important when only few examples are available, which is typically the case in practice.
For instance, in the data set used in our empirical study $42.61\%$ of road segments have been traversed ten times or less over an 18 month period.

The paper's overall contribution is \glsfirst{aha}, a Bayesian travel time and speed estimation framework that integrates (and complements) function-fitting and aggregation-based approaches to leverage the strengths of both.
In addition, we present an instance of the \gls{aha} framework, \gls{agha}, that models travel time or speed as a Gaussian variable with unknown mean and variance.
The use of conjugate priors in \gls{agha} allows for efficient computation of the posterior, easy implementation, and makes \gls{agha} applicable to neural network learning using standard deep learning frameworks.
Finally, we investigate the capabilities of the \gls{aha} framework in an empirical study using \gls{agha}.
\todo{update improvement numbers}
The study shows that \gls{aha} can achieve $12.93\%\text{-}32.18\%$ and $8.34\%\text{-}18.55\%$ better performance in terms of travel speed distribution modeling and travel time point estimation, respectively, compared to using a function-fitting or aggregation-based approach alone.
In addition, \gls{aha} can achieve better generalizability than function-fitting approaches while maintaining similar or better accuracy to aggregation-based approaches regardless of data availability.

The remainder of the paper is structured as follows.
\cref{sec:preliminaries-paper5} provides the necessary background on function-fitting and aggregation-based approaches, as well as graph modeling of road networks.
\cref{sec:aha-paper5} presents the \gls{aha} framework and describe how \gls{aha} can unify existing function-fitting and aggregation-based approaches.
\cref{sec:aha-gaussian-paper5} presents the Gaussian instance of the \gls{aha} framework, \gls{agha}.
\cref{sec:experiments-paper5} reports on the empirical study.
\cref{sec:related-work-paper5} reviews related work, and \cref{sec:epilogue-paper5} concludes and offers directions for future research.


\section{Preliminaries}\label{sec:preliminaries-paper5}
We now provide the necessary background on data modeling and existing approaches to travel time and speed estimation.

\subsection{Data Modeling}
For clarity, we present \gls{aha} as well as existing function-fitting and aggregation-based approaches for routes, i.e., where travel times and travel speeds are estimated for routes in in a road network with road segments as a special case.
\gls{aha} is applicable to other kinds of data as well, to be discussed in \cref{sec:related-work-paper5}.

\subsubsection{Road Network Modeling}\label{sec:road-network-modeling-paper5}
A road network is modeled as a directed graph $G=(V, E)$ where a vertex $v \in V$ represents an intersection or the end of a road, and an edge $e \in E$ represents a road segment.
A route is a connected path $p = (e_1, \dots, e_n)$ where $e_i \in E$ for $1 \leq i \leq n$.
In addition, a route can be mapped to a $d$-dimensional feature vector describing its characteristics using the mapping function $\phi$.
For brevity, we use the notation $\mathbf{p}$ to refer to the feature vector representation $\phi(p)$ of a route $p$.
If $p$ consists of one edge, i.e., $p=e$, we use the notation $\mathbf{e}$ instead.

\subsubsection{Trajectory Modeling}\label{sec:trajectory-modeling-paper5}
Vehicle trajectories are sequences of time-stamped \gls{gps} locations, but can be map-matched to a road network modeled as a directed graph (as described in \cref{sec:road-network-modeling-paper5}).
Each map-matched trajectory is a sequence $\mathit{TR}=(\mathit{tr}_1, \dots, \mathit{tr}_n)$ where $\mathit{tr}_i=(e_i, \tau_i, t_i)$ is a triple consisting of a road segment $e_i \in E$, an arrival time $\tau_i$ corresponding to the timestamp of the first recorded \gls{gps} location on road segment $e_i$, and $t_i$ the travel time or travel speed recorded during the traversal of segment $e_i$.
In some cases, the traversal of a road segment in a trip is inferred by the map-matching algorithm due to a lack of \gls{gps} data.
In such cases, $t_i = \varnothing$ and $\tau_i = \varnothing$.

\subsection{Existing Approaches}
We now describe function-fitting and aggregation-based approaches to travel time or speed estimation for routes.

\subsubsection{Function-Fitting Approaches}
Function-fitting approaches%
~\citep{wei2020spatial,lu2020st,ge2020global,zhang2020novel,zhang2020graph,zhang2020network, yin2020attention,lee2020predicting,guo2019attention,cui2019traffic,yu2017spatio,htte,stad,zheng2013time,fu2020estimation,barnes2020bustr,lan2019travel,wu2019deepeta,shen2019tcl,hu2019stochastic,hu2020stochastic,taoyang2019deepist,xi2019path,compacteta,rade2018wedge,zheng2018learning,wang2018will,wang2014travel,yang2013using,workshop}
assume that the relationship between the unknown future travel time or travel speed $\hat{t}_i$ when traversing route $p_i$ at time $\tau_i$ can be modeled by a function $f$ s.t. $\Pr(\hat{t}_i \mid p_i, \tau_i; \psi) = f(\mathbf{p}_i, \boldsymbol\tau_i; \psi)$ where $\psi$ is the function parameters of $f$, $\mathbf{p}_i$ is the feature vector representation $\mathbf{p}_i$ of route $p_i$, and $\boldsymbol\tau_i$ is the vector representation of time $\tau_i$.

As an example of a function $f$, if the probability density $\Pr(\hat{t}_i \mid p_i, \tau_i; \psi)$ is a uni-variate Gaussian, then a possible choice of $f$ is $f(\mathbf{p}_i, \boldsymbol\tau_i; \psi) = \mathcal{N}(\hat{t}_i \mid \boldsymbol\psi_{\mu}(\mathbf{p}_i \oplus \boldsymbol\tau_i), \boldsymbol\psi_{\sigma^2}(\mathbf{p}_i \oplus \boldsymbol\tau_i))$ where $\oplus$ denotes vector concatenation and the parameters $\psi=\{\boldsymbol{\psi}_{\mu}, \boldsymbol{\psi}_{\sigma^2}\}$ consists of two real vectors.
In this case, the parameters $\boldsymbol{\psi}_{\mu}$ and $\boldsymbol{\psi}_{\sigma^2}$ that are used to compute the mean and variance of the Gaussian, respectively.

To fit a function $f$ to a set of $n$ training trajectories, function-fitting approaches learn model parameters $\theta$ that maximize the conditional likelihood
$
  \prod_{i=1}^n \Pr(\hat{t}_i \mid p_i, \tau_i; \psi) = f(\mathbf{p}_i, \boldsymbol\tau_i; \psi).
$
By sharing model parameters $\psi$ across training trajectories during optimization, function-fitting approaches can generalize to unseen routes. However, in practice, generalization is imperfect for non-trivial travel time and speed estimation tasks.

\subsubsection{Aggregation-Based Approaches}
Aggregation-based approaches%
~\citep{pace,hu2017enabling,dai2016path,yuan2011tdrive}
aim to learn model parameters $\theta_i$ for each route at time $\tau_i$ using a set of training trajectories.
In other words, given a set of $n$ training trajectories, these approaches solve $n$ optimization problems.

Unlike function-fitting approaches, aggregation-based approaches do not optimize the posterior predictive directly.
Instead, the model parameters $\theta_i$ are chosen s.t.\ they are the maximum a posteriori probability (MAP) estimate of
$
  \Pr(\theta_i \mid \tau_i, \tilde{T}_i)
$,
where $\tilde{T}_i$ is a set of historical travel speed or time records that are typically collected from historical trajectories traversing route $p_i$ during some interval based on time $\tau_i$, e.g., the same time-of-week interval as $\tau_i$.

By using distinct model parameters for each route set of training trajectories, aggregation-based approaches can learn a more accurate representation of the travel time or travel speed distribution of $\hat{t}_i$ provided sufficient historical records are available.
However, in practice, it is unlikely that there is sufficient data for all routes at all times of day~\citep{yang2013using,tan2020cycle,wei2020a}.
In such cases, aggregation-based approaches rely on simplistic heuristics to provide reasonable estimates for unseen routes, and they are prone to overfitting when only a few historical records are available.

\section{A Unified Approach}\label{sec:aha-paper5}
We now present the proposed \gls{aha} framework.


\subsection{Framework}
The primary goal of \gls{aha} is to unify a function-fitting component with an aggregation-based component s.t.\ we can seamlessly and smoothly switch between the two to leverage their respective strengths.
The two components should be integrated s.t.\ \gls{aha} relies on the function-fitting component when no historical data is available or if the data is not representative, e.g., due to low availability.
Conversely, we want \gls{aha} to rely on the aggregation-based component when historical data abounds.
To achieve this, \gls{aha} adopts a Bayesian foundation that provides a solid theoretical basis for unifying function fitting and aggregation.

\begin{figure}[h]
  \centering
  \usetikzlibrary{bayesnet,arrows,positioning,shapes,arrows.meta}
\begin{tikzpicture}[thick,
                    parameter/.style={rectangle, black, draw},
                    minimum width=0.7cm,
                    align=center,
                    arr/.style = {thick, ->}]
  \node[latent] (D) {$\hat{t}_i$};
 
  \node[latent, below=1cm of D] (paramsi) {$\theta_i$};
  \node[obs, right=1cm of paramsi] (Dtilde) {$\tilde{t}_{i, j}$};
  \plate[] {Dtildeplate} {(Dtilde)} {$j=1:m$};

  \edge {paramsi} {D};
  \edge {paramsi} {Dtilde};
 
  \node[parameter, below=1cm of paramsi] (function) {$f(\mathbf{p}_i, \boldsymbol\tau_i; \psi)$};
  \edge {function} {paramsi};
  \node[parameter, left=1cm of function] (theta) {$\psi$};
  \edge {theta} {function};
  \node[parameter, yshift=0.3cm, right=0.56cm of function] (route) {$p_i$};
  \edge {route} {function};
  \node[parameter, yshift=-0.3cm,right=0.56cm of function] (time) {$\tau_i$};
  \edge {time} {function};
  \coordinate[left=0.275cm of function] (plateexpand1) {};
  \coordinate[right=0.275cm of time] (plateexpand2) {};
  \plate[]{componentplate} {(D)(Dtildeplate)(function)(route)(time)(plateexpand1)(plateexpand2)} {$i=1:n$};

\end{tikzpicture}
  \caption{The \gls{aha} framework illustrated using plate notation.\label{fig:conceptual-framework}}
\end{figure}
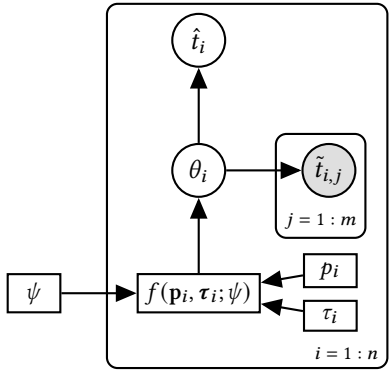

A conceptual model of \gls{aha} is illustrated in \cref{fig:conceptual-framework}.
In brief, we assume that the travel time or speed distribution of a route $p_i$ at time $\tau_i$ follows a distribution with uncertain hyperparameters $\theta_i$.
The prior distribution of $\theta_i$ is computed as $\Pr(\theta_i \mid f(\mathbf{p}_i, \boldsymbol\tau_i; \psi))$ using a \emph{prior function} $f$ with function parameters $\psi$ that are shared across all routes.
The prior function $f$ represents the function-fitting component of \gls{aha}, and allows \gls{aha} to estimate prior distributions for routes even if no historical data is available at estimation time.
This estimate is based on the travel time or travel speed distributions of similar routes at similar times.
However, if historical records $\tilde{T}_i = \{\tilde{t}_{i, 1}, \dots, \tilde{t}_{i, m}\}$ are available, a posterior travel time or speed distribution $\Pr(t_i \mid \tilde{T}_i;\psi)$ for route $p_i$ at time $\tau_i$ is computed.
The computation of the posterior represents the aggregation-based component in \gls{aha}.

\subsection{The \gls{aha} Objective}\label{sec:hybrid-objective}
We now present the objective function used to train models within the \gls{aha} framework.

\subsubsection{Objective Function}
\cref{fig:conceptual-framework} depicts a generative model, i.e., a model that specifies how to generate the new records from the distribution $\Pr(\hat{t}_i \mid \theta_i)$~\citep{murphy2012machine}.
Generative models are usually trained by selecting parameters $\theta_i$ that maximize the joint likelihood~\citep{murphy2012machine}.
However, training a generative model by maximizing the conditional likelihood is guaranteed to yield better estimations when the true travel time or travel speed distribution is different from the distribution family assumed by the model~\citep{salojarvi2005discriminative}.
This is generally the case in practice, where, e.g., travel time distributions are highly complex~\citep{pace}.
In this work, we are interested in predictive performance and therefore maximize the conditional likelihood
\begin{equation}\label{eq:hybrid-objective}
  \Pr(t_i \mid \tilde{T}_i, \theta_i) = 
  \Pr(t_i \mid \tilde{T}_i, \mathbf{p}_i, \boldsymbol\tau_i; \psi)
\end{equation}
across $n$ training trajectories where $t_i$ is the ground truth travel time or travel speed observed in the $i$th trajectory when traversing route $p_i$ at time $\tau_i$.
The posterior predictive $\Pr(\hat{t}_i \mid \tilde{T}_i, \theta_i)$ equals the prior predictive $\Pr(\hat{t}_i \mid \theta_i)$ if no historical records are available as evidence, i.e., if $\tilde{T}_i = \emptyset$.

\subsubsection{Regularizing Properties}\label{sec:regularizing-properties}
The \gls{aha} framework inherently addresses the issues of data imbalance issues of function-fitting approaches where they tend to fit best to frequently occurring types of, e.g., road segments.
Because the \gls{aha} framework is Bayesian, the \gls{aha} objective in \cref{eq:hybrid-objective} is implicitly regularized s.t.\ the performance of the function-fitting component represented by prior function $f$ is inversely proportional to the number of historical records available.
In other words, the function-fitting component is trained to perform well in data-sparse situations.

The posterior predictive in \cref{eq:hybrid-objective}, i.e.,
\begin{equation*}
  \Pr(\hat{t}_i \mid \tilde{T}_i, \theta_i) = \int_{\theta_i} \Pr(\hat{t}_i \mid \theta_i)\Pr(\theta_i \mid \tilde{T}_i) \mathit{d\theta_i}\text{,}
\end{equation*}
depends on the posterior distribution
\begin{equation}\label{eq:regularization-properties}
  \Pr(\theta_i \mid \tilde{T}_i) \propto \Pr(\theta_i) \prod_{j=1}^m \Pr(\tilde{t}_{i,j} \mid \theta_i).
\end{equation}
As \cref{eq:regularization-properties} shows, the importance of the prior distribution $\Pr(\theta_i)$ on the posterior distribution, and thus the posterior predictive, is inversely proportional to the number of historical records.
As a consequence, the influence of the function-fitting component on the \gls{aha} objective in \cref{eq:hybrid-objective} is largest when there are no historical records at all, and it gradually becomes less important if more historical records are available.
Thus, by maximizing the conditional likelihood in \cref{eq:hybrid-objective}, the function-fitting component is trained to perform well in data-sparse situations.
This ensures that no explicit regularization of the function-fitting component, e.g., oversampling~\citep{workshop}, is required to handle data imbalance issues.

\subsection{Relation to Existing Approaches}
\gls{aha} can be viewed as a hybrid of function-fitting and aggregation-based approaches.

\subsubsection{Hybrid Characteristics}
The \gls{aha} framework and it corresponding objective in \cref{eq:hybrid-objective} are hybrid in the sense that, like the function-fitting approaches, it fits a function $f$ that maps input feature vector representations to a prior travel time or travel speed estimate by minimizing the discrepancy between the output of the mapping and the expected output.
However, like the aggregation-based approaches, \gls{aha} can also use historical records directly in the estimation process, i.e., without a typically imperfect intermediary mapping function, to adjust the prior estimate of its function-fitting component by computing the posterior.
This adjustment allows \gls{aha}, like aggregation-based approaches, to approximate a travel time or travel speed distribution at arbitrary precision given sufficient data and an appropriate choice of the distribution family for $\Pr(\hat{t}_i \mid \tilde{T}_i, \mathbf{p}_i; \theta)$.

A very appealing property of \gls{aha} is that from a modeling capability perspective, \gls{aha} models are capable of being at least as powerful as either their function-fitting or aggregation-based components.
Specifically, a \gls{aha} model can match the performance of its function-fitting component at arbitrary precision by expressing very high confidence in the prior.
Similarly, a \gls{aha} model can match the performance of its aggregation-based component at arbitrary precision by expressing very low confidence in the prior.

\subsubsection{Integration with Existing Approaches}\label{sec:integration}
An important feature of the \gls{aha} framework is that it is complimentary and integrable with existing approaches to travel time and speed estimation.
Within the framework, existing function-fitting approaches provide the structure of the prior function $f$ used to estimate the prior hyperparameters.
We expect that most function-fitting approaches can be integrated with the \gls{aha} framework with only minor modifications to the output layer and the objective function, depending on the choice of distribution for $\hat{t}$.
Next, existing aggregation-based approaches are primarily concerned with the selection of historical records for aggregation.
Aggregation-based approaches thus provide record selection strategies to construct the set of historical records $\tilde{T}_i$ in \cref{eq:hybrid-objective}.

\subsubsection{Remarks on Training}
Unlike aggregation-based approaches, \gls{aha} maximizes the conditional likelihood, and it is desirable that a function-fitting component (represented by prior function $f$) compensates for an aggregation-based component when insufficient historical data is available.
In the extreme case, no data may be available.
To simulate this situation during training, we recommend excluding the ground truth travel time or travel speed from the set of historical records, i.e., we recommend that $t_i \notin \tilde{T}_i$ in \cref{eq:hybrid-objective}. 

By excluding the ground truth travel time or travel speed from the set of historical records during the training, the function-fitting component must always contribute information missing from the set of historical records $\tilde{T}_i$ during training.
Specifically, it must contribute information about the ground truth travel time or speed $t_i$ to optimize the objective in \cref{eq:hybrid-objective}.

\section{{Gaussian UniTE}}\label{sec:aha-gaussian-paper5}
\gls{aha} is a framework for which many instantiations are possible.
To study the prospects of \gls{aha} analytically and empirically, we present one such instantiation, \glsfirst{agha}, that is both easy to implement and integrate with existing approaches. 

As the name suggests, \gls{agha} assumes that $\hat{t}_i \mid \theta_i$ follows a Gaussian distribution with parameters $\theta_i =(\mu_i, \lambda_i)$.
The Gaussian assumption combined with the use of conjugate priors over the uncertain mean $\mu_i$ and precision $\lambda_i$ allows the generally difficult-to-compute posterior predictive in \cref{eq:hybrid-objective} to be computed efficiently and in closed-form~\citep{murphy2007conjugate}.
In addition, the closed-form computation of the posterior predictive is also differentiable.
This enables the use of gradient-based optimization techniques that are commonly used in function-fitting approaches based on neural networks.

Note that \gls{aha} is far more general than \gls{agha} which is just one possible instantiation of the \gls{aha} framework.
Differentiability of the posterior predictive is a convenient property of \gls{agha} but the \gls{aha} framework is not restricted to gradient-based optimization.

\subsection{Prior}\label{sec:agha-prior}
Let $\Pr(\hat{t}_i \mid \mu_i, \lambda_i)$ denote the likelihood of a Gaussian distribution with mean $\mu_i$ and variance $\sigma_i^2 = \lambda^{-\frac{1}{2}}$.
We adapt the work of \citet{murphy2007conjugate} to our setting, and estimate $\mu_i$ and $\lambda_i$ using the normal-gamma prior
\begin{multline}\label{eq:conjugate-priors}
    \Pr(\mu_i, \lambda_i \mid \mathbf{p}_i; \psi) = \mathit{NG}(\mu_i, \lambda_i \mid \mu_{i,0}, \kappa_{i,0}, \alpha_{i,0}, \beta_{i, 0}) = \\
    \mathcal{N}(\mu_i \mid \mu_{i, 0}, \frac{1}{\kappa_{i,0} \lambda_{i, 0}})\mathit{Ga}(\lambda_i \mid \alpha_{i, 0}, \beta_{i, 0}),
\end{multline}
where $f(\mathbf{p}_i; \psi) = \begin{bmatrix}\mu_{i,0} & \kappa_{i,0} & \alpha_{i,0} & \beta_{i, 0}\end{bmatrix}$.
Here, $\kappa_{i, 0} > 0$, shape parameter $\alpha_0 > 0$, and rate parameter $\beta_0 > 0$.
In other words, the parameters of the prior, or the \emph{prior hyperparameters}, are output by the function $f$.

\subsection{Posterior}\label{sec:agha-posterior}
After observing a sample of $m$ historical records $\tilde{T}_i = \{\tilde{t}_{i, 1}, \dots, \tilde{t}_{i, m}\}$, beliefs about $\mu$ and $\lambda$ may change.
Formally, the posterior distribution over $\mu_i$ and $\lambda_i$ is given as ~\citep{murphy2007conjugate}
\begin{multline}\label{eq:conjugate-posteriors}
    \Pr(\mu_i, \lambda_i \mid \mathbf{p}_i, \tilde{T}_i; \theta) \mathit{NG}(\mu_i, \lambda_i \mid \mu_{i,m}, \kappa_{i,m}, \alpha_{i,m}, \beta_{i, m}) = \\
   \mathcal{N}(\mu_i \mid \mu_{i, m}, \frac{1}{\kappa_{i,m} \lambda_{i, m}})\mathit{Ga}(\lambda_i \mid \alpha_{i, m}, \beta_{i, m}),
\end{multline}
with \emph{posterior hyperparameters}
\begin{equation}\label{eq:posterior-parameters}
\begin{split}
  \mu_{i, m}    =& \frac{\kappa_{i, 0}\mu_{i, 0} + mM_{\tilde{T}_i}}
                   {\kappa_{i, 0} + m} \\
  \kappa_{i, m} =& \kappa_{i, 0} + m \\
  \alpha_{i, m} =& \alpha_{i, 0} + \frac{m}{2}\\
  \beta_{i, m} =& \beta_{i, 0} +
             \frac{1}{2}mS^2_{\tilde{T}_i} +
             \frac{1}{2}\frac{\kappa_{i, 0}m{(M_{\tilde{T}_i} - \mu_{i, 0})}^2}
                             {\kappa_{i, 0} + m},
\end{split}
\end{equation}
where $f(\mathbf{p}_i; \theta) = \begin{bmatrix}\mu_{i,0} & \kappa_{i,0} & \alpha_{i,0} & \beta_{i, 0}\end{bmatrix}$, $M_{\tilde{T}_i} = \frac{\sum_{j=1}^n \tilde{T}_{i, j}}{m}$ is the sample mean, and $S^2_{\tilde{T}_i} = \frac{\sum_{j=1}^m{(\tilde{t}_{i, j} - M_{\tilde{T}_i})}^2}{m}$ is the biased sample variance.
Note that if there is no data, i.e., $m = 0$, then the posterior hyperparameters are equal to the prior hyperparameters.

The regularizing properties of the \gls{aha} framework discussed in \cref{sec:regularizing-properties} are reflected in the formulas for the posterior hyperparameters in \cref{eq:posterior-parameters}.
For instance, the posterior mean $\mu_{i, m}$ is a weighted mean of the prior mean $\mu_{i, 0}$ and the mean of the historical records $M_{\tilde{T}_i}$ where $\mu_{i, 0}$ has weight $\kappa_{i, 0}$ and $M_{\tilde{T}_i}$ has weight $m$.
Thus, the influence of the prior mean $\mu_{i, 0}$ on the posterior mean $m_{i, m}$ diminishes as $m$ increases.
The remaining posterior hyperparameters follow the same pattern.

\subsection{Posterior Predictive}
It follows from the posterior in \cref{eq:conjugate-posteriors}, that the posterior predictive $\Pr(\hat{t}_i \mid \mathbf{p}_i \tilde{T}_i; \theta)$---which we seek to optimize in the objective function in \cref{eq:hybrid-objective}---follows a student's $t$-distribution
$t_{\nu_i}(\hat{t}_{i} \mid \hat{\mu}_i, \hat{\sigma}_i)$
with $\nu_i = 2\alpha_{i, m}$ degrees of freedom, location $\hat{\mu}_i = \mu_{i, m}$, and scale $\hat{\sigma}_i = \sqrt{\frac{\beta_{i, m}(\kappa_{i, m} + 1)}{\alpha_{i, m}\kappa_{i, m}}}$~\citep{murphy2007conjugate}, and with probability density function
\begin{multline}\label{eq:gaussian-posterior-predictive}
  h(t_i \mid \nu_i, \hat{\mu}_i, \hat{\sigma}_i) = 
    \frac{\Gamma(\frac{\nu_i +1}{2})}
         {\Gamma(\frac{\nu_i}{2}) \sqrt{\nu_i \mathrm{\pi}} \hat{\sigma}_i }  
    {\Bigg(
        1 + \frac{1}{\nu_i}  
        {\Big( \frac{t_i - \hat{\mu}_i}{\hat{\sigma}_i} \Big)}^2 
    \Bigg)}^{-\frac{\nu_i+1}{2}}.
\end{multline}

\subsection{A Prior Function Layer}
To illustrate how to use \gls{agha} with neural networks, we present a prior function layer in \cref{alg:hybrid-output-layer} that outputs the prior hyperparameters in \gls{agha}.
The prior function layer is intended to be used as the final layer of a neural network s.t.\ the neural network models the prior function $f(\mathbf{p}_i, \boldsymbol\tau_i; \psi)$ in \cref{fig:conceptual-framework}, where $\psi$ are neural network weights.

\begin{algorithm}
  \caption{Forward Propagation through the Prior Function Layer\label{alg:hybrid-output-layer}}
  \begin{algorithmic}[1]\raggedright
    \Function{PriorFunctionLayer}{$\mathbf{x}_i$}
        \State{$x_i \gets h(\mathbf{p}_i, \boldsymbol\tau_i; \psi_h)$}
        \State{$\mathbf{h}_1 \gets \mathbf{W} \cdot \mathbf{p}_i$}
        \State{\Let{$\mathbf{h}_1 = \begin{bmatrix} h_{1,1} & h_{1,2} & h_{1,3} & h_{1,4}\end{bmatrix}$}}
          \State{$\mu_{i, 0} \gets h_{1,1}$}
          \State{$\kappa_{i, 0} \gets \textsc{ELU}_{a}(h_{1,2}) + a + \epsilon$}
          \State{$\alpha_{i, 0} \gets \lvert h_{1,3} \rvert + \epsilon$}
          \State{$\beta_{i, 0} \gets \lvert h_{1,4} \rvert + \epsilon$}
          \State{\Return{$\begin{bmatrix} \mu_{i,0} & \kappa_{i, 0} & \alpha_{i,0} & \beta_{i, 0}\end{bmatrix}$}}
    \EndFunction
  \end{algorithmic}
\end{algorithm}

The prior function layer in \cref{alg:hybrid-output-layer} takes as input a feature vector $\mathbf{x}_i$.
In the context of neural networks, $\mathbf{x}_i$ may be the result of a function $h$ s.t.\ $h(\mathbf{p}_i, \boldsymbol\tau_i) = \mathbf{x}_i$ where $h$ represents forward propagation of vectors $\mathbf{p}_i$ and $\boldsymbol\tau_i$ through multiple layers.
In lines 3--4, $\mathbf{x}_i$ is projected to a four-dimensional vector $\mathbf{h}_1$, one for each of the prior hyperparameters, using a learnable weight matrix $\mathbf{W}$.
Recall from \cref{sec:agha-prior} that the prior hyperparameters are constrained s.t.\ $\kappa_{i, 0} > 0$, $\alpha_0 > 0$, and $\beta_0 > 0$.
These constraints are enforced in lines 6--8.
The values $h_{1,3}$ and $h_{1,4}$ are interpreted as the prior hyperparameters $\alpha_{i,0}$ and $\beta_{i,0}$ and are constrained by taking their absolute values and adding a small non-zero positive constant $\epsilon$ to ensure that they are greater than zero.
We chose this way of enforcing non-negativity due to its simplicity.

We initially constrained the value $h_{1,2}$, interpreted as the prior hyperparameter $\kappa_{i, 0}$, in the same way as $h_{1, 3}$ and $h_{1, 4}$.
However, as discussed in \cref{sec:agha-posterior}, $\kappa_{i, 0}$ represents the confidence in the prior, i.e., the output of the function-fitting component.
Experiments showed that, since the function-fitting component performs poorly in the initial stages of training, the value of $\kappa_{i, 0}$ will be very low.
To alleviate this problem, we use the expression in line $6$ of \cref{alg:hybrid-output-layer} instead, which makes use of the \gls{elu}~\citep{elu} function 
\begin{equation}\label{eq:elu}
  \text{ELU}_a(x) = 
  \begin{cases}
    x & x > 0 \\
    a(\mathrm{e}^x-1) & x \leq 0,,
  \end{cases}
\end{equation}
where $a > 0$.

Because \cref{eq:elu} has a minimum value of $-a$, we can enforce non-negativity of $\kappa_{i, 0}$ by adding $a$ and $\epsilon$, as shown in line $6$ of \cref{alg:hybrid-output-layer}.
This expression makes the value of $\kappa_{i, 0}$ less sensitive to changes that decrease its value, thus discouraging decreases of $\kappa_{i, 0}$ during early stages of training that needs to be corrected in later stages.
Hyperparameter $a$ regulates this effect, s.t.\ the effect is inversely proportional to $a$.
In addition, the value of $h_{1,2}$ is initially very close to zero in a neural network setting. 
Using $\kappa_{i, 0} = |h_{1,2}| + \epsilon$ would therefore result in a $\kappa_{i, 0}$ value close to zero indicating, an unreasonably low confidence in the model.
The constraint measure used in line~6 in \cref{alg:hybrid-output-layer} instead ensures that the initial value of $\kappa_{i,0}$ is close to $a$.
Preliminary experiments showed performance improvements when enforcing non-negativity of $\kappa_{0, i}$ in this way, as opposed, taking the absolute value and adding a small constant, but they showed improvements when non-negativity of $\alpha_{i, 0}$ and $\beta_{i, 0}$ were enforced in the same way.

\section{Empirical Study}\label{sec:experiments-paper5}
We evaluate \gls{aha} on the task of trajectory travel time estimation.
In particular, we are interested in evaluating \gls{aha}'s capability for improving estimation of the travel speed distributions of road segments traversed during a trip over function-fitting and aggregation-based approaches, but also \gls{aha}'s capability for improving point estimates of travel times.
In addition, we investigate the behavior of \gls{aha} under varying degrees of data availability and different choices of parameters.

\subsection{Dataset}\label{sec:data-set-paper5}
For our experiments, we use a dataset of $336\,253$ trajectories from Aalborg Municipality in Denmark that occurs between January 1st 2012 and December 31st 2014~\citep{trajectorydata}.
The trajectories have been map-matched to the road network of Aalborg Municipality extracted from \gls{osm}~\citep{osm} with $16\,294$ intersections and $35\,947$ road segments.
See \cref{sec:road-network-modeling-paper5} and \cref{sec:trajectory-modeling-paper5} for details on road network trajectory modeling, respectively.
See~\citep{trajectorydata} for details on the trajectory data and map-matching process.

We use the $148\,746$ trajectories from the period January 1st 2012 to June 31st 2013 for training and set aside the $72\,693$ trajectories from July 1st 2013 to December 31st 2013 for validation.
We use the remaining $114\,028$ trajectories from January 1st 2014 to December 31st 2014 for testing.
To characterize each road segment in a trajectory, we use a set of $16$ features derived from \gls{osm} data and data from the Danish business authority~\citep{rfnlong}. These road segment feature representations are sparse, containing information about just four attributes: road segment length, road segment category (e.g., motorway), and the kind of zone (city, rural, or summer cottage) the source and target intersections of the road segment are in.
The sparsity in the feature representation makes function-fitting difficult~\citep{workshop,rfnlong}.
In addition, $19\,510$ ($54\%$) of the road segments are annotated with a speed limit derived from \gls{osm} and municipal data~\citep{rfnlong}.
For further details, see \citep{rfnlong}.

\COMMENT{
\subsection{Representation of Time}\label{sec:representation-of-time}
Travel speed estimation is a time-dependent estimation task.
We therefore learn a $16$-dimensional feature vector representation of the time of week, where $8$ dimensions are used to represent the time of day and $8$ dimensions are used to represent the day of the week.
To represent the time of day, we divide the time of day into $96$ $15$-minute intervals and learn an $8$-dimensional vector for each time interval.
Preliminary experiments indicated that our results are robust to changes to the dimensionality of this feature vector representation.
We refer to \cref{app:representation-of-time} for details on the time of week vector representation.
}

\subsection{Objective Function}
We optimize for travel speed distribution modeling performance using the per-trajectory mean \gls{nll}.
The \gls{nll} of a trajectory $\mathit{TR}$ is
\begin{equation}\label{eq:nll}
  \text{NLL}(\mathit{TR}) = \sum_{(e_i, \tau_i, t_i) \in \mathit{TR}, t_i \neq \varnothing} \text{sNLL}(e_i, \tau_i, t_i)
\end{equation}
where $\text{sNLL}(e_i, \tau_i, t_i) = -\ln \Pr(t_{i} \mid \theta)$ and $\theta$ are model parameters. 
In the case of \gls{aha}, $\Pr(t_{i} \mid \theta)$ is the Gaussian posterior predictive (see \cref{eq:gaussian-posterior-predictive}).
The \gls{nll} directly measures the likelihood of a trajectory occurring by considering how well an algorithm models the travel speed distributions of its constituent road segments.
A good model achieves a high likelihood across all trajectories, resulting in a low \gls{nll} score.

\subsection{Algorithms}\label{sec:exp-algos}
In our empirical study, we combine a function-fitting baseline and an aggregation-based baseline in a unified approach using the \gls{aha} framework and compare their separate performance with their unified performance.

The output of all algorithms is the density function $\Pr(\hat{t}_i=t_i \mid \theta)$ used in \cref{eq:nll} where $\theta$ are model parameters specific to each algorithm.
All algorithms are implemented in Python~3\footnote{https://www.python.org/} and trained using the MXNet deep learning framework\footnote{\url{https://mxnet.incubator.apache.org}}.
We have made the implementation of all algorithms publicly available\footnote{To be released upon acceptance.}.

\subsubsection{AGG}
Existing aggregation-based approaches~\citep{pace,hu2017enabling,dai2016path,yuan2011tdrive} are very similar and we use the AGG baseline to represent these approaches in our empirical study. 

Firstly, when estimating the travel speed distribution of a road segment $e$ at time $\tau$, these approaches aggregate historical records from trajectories where the road segment is traversed at a similar time within some interval of size $\delta$.
Secondly, rather than modeling uncertainty about the hyperparameters of the distribution model like \gls{aha}, they set a threshold $k$ for the minimal number of historical records considered sufficient.
If the number of historical records is insufficient, an estimate is derived from the speed limit~\citep{pace,hu2017enabling,dai2016path,yuan2011tdrive}.

We represent existing aggregation-based approaches using a single baseline algorithm AGG with the two features of existing aggregation-based approaches.
For a fair comparison with \gls{agha}, AGG also models travel speed distributions as Gaussian distributions rather than histograms~\citep{pace,hu2017enabling,dai2016path} or a mean value~\citep{yuan2011tdrive}.
See \cref{app:aggregation-baseline-paper5} for a detailed description of the AGG baseline.

\paragraph{Record Selection}
AGG's performance is strongly dependent on the record selection strategy used.
In general, aggregation-based approaches must balance \emph{record relevance} with \emph{record availability}
The selection strategy used by AGG considers two kinds of relevance: contextual relevance and temporal relevance.

The contextual relevance hyperparameter $c$ is an integer that adjusts the contextual relevance where the context is the $c$ preceding and succeeding road segments in the trajectory from which a historical records originates.
Only historical travel speed records from the training trajectories with the same context are selected for aggregation.
Thus, increasing $c$ increases contextual relevance.

The temporal relevance parameter $\delta$ is given in some unit of time and adjusts how inclusive the record selection strategy is w.r.t.\ historical records from different times of week.
For instance, when estimating for time $\tau$, contextually-relevant historical records occurring in the time interval $[\tau_i - \frac{\delta}{2}; \tau_i + \frac{\delta}{2}]$ are selected.

The record selection algorithm is described fully in \cref{app:record-selection}.

\subsubsection{GRU}
Recurrent neural networks are a popular neural network architecture among function-fitting literature.
We therefore use a simple recurrent neural network, \emph{GRU}, as the function-fitting approach in our study, which features a GRU cell~\citep{gru} with a skip connection.
The GRU model uses the prior function layer described in \cref{alg:hybrid-output-layer} as the final layer.
GRU is fully specified in \cref{app:gru}.

Unlike AGG, recurrent models can model correlations in segment travel speeds within a trajectory.
For instance, if a vehicle drives onto a motorway segment `A' from a motorway approach `B' with a lower speed limit than the motorway, some time is spent on acceleration to cruising speed.
On the other hand, if the vehicle drove onto motorway segment `A' from an adjacent motorway segment `C' less time, if any, is spent on acceleration assuming similar traffic conditions.

\subsubsection{\gls{aha}}
We unify the AGG and GRU baselines using the \gls{aha} framework following the instructions in \cref{sec:integration}.
In brief, we use the function-fitting approach, GRU, to estimate the prior and the record selection strategy of the aggregation-based approach, AGG, to compute the posterior.

As mentioned in \cref{sec:hybrid-objective}, we train the generative \gls{aha} model using a discriminative objective.
To evaluate the this decision, we consider both a discriminative and a generative variant of \gls{aha}. 

UniTE-DIS is UniTE as described in \cref{sec:aha-paper5} where the posterior predictive is optimized directly during training.
This is considered an \emph{end-to-end} approach from a machine learning perspective.
UniTE-GEN instead operates in two step.
First, UniTE-GEN outputs only the prior predictive at training time and then computes the posterior predictive only at test time.

A benefit of UniTE-GEN is that it can be applied to already training function-fitting approaches.
In our empirical study, we always reuse a GRU model for UniTE-GEN in a one-to-one fashion.
That is, whenever we train and evaluate a GRU model, we also evaluate the corresponding UniTE-GEN model.

\subsection{Evaluation Metrics}
We use \gls{nll} (see \cref{eq:nll}) to evaluate travel speed distribution estimation of each algorithm in our study.
In addition, we evaluate each algorithm's travel time point estimation performance using \gls{mae}, a commonly used measure for this purpose.
Finally, to make our results more easily comparable to those of other papers using different methods and datasets, we also measure the \gls{mape} of the algorithms used in our empirical study, another commonly used measure in the traffic travel time and speed estimation literature.

We compute a travel time point estimate for a trajectory following route $p = (e_1, \dots, e_n)$ as 
$
   \sum_{i=1}^n \frac{l_i}{\mathbb{E}[d_i]} = \sum_{i=1}^q \frac{l_i}{\hat{\mu_{i}}}
$
where $l_i$ is the length of road segment $e_i$ and $\hat{\mu}_i = \mu_{i, m}$ is the expected travel speed computed using \cref{eq:posterior-parameters}. 
Recall that $m=0$ for GRU.

The \gls{mae} is the mean absolute error of the estimated and actual travel time point estimate.
For a trajectory $\mathit{TR}$, the absolute error is
  $
  \text{AE} = |\hat{y}_{\mathit{TR}} - y_{\mathit{TR}}|
  $
where $\hat{y}_{\mathit{TR}}$ and $y_{\mathit{TR}}$ is the travel time point estimate and ground truth, respectively, of trajectory $T$.

The \gls{mape} is the mean absolute percentage error of the estimated and actual travel time point estimate.
For a trajectory $\mathit{TR}$, the absolute percentage error is
  $
  \text{APE} = \frac{|\hat{y}_{\mathit{TR}} - y_{\mathit{TR}}|}{y_{\mathit{TR}}}.
  $

\subsection{Training and Hyperparameter Selection}
The GRU and UniTE-DIS models are trained by to minimizing the \gls{nll} in \cref{eq:nll} across all trajectories in the training using the ADAM optimizer~\citep{adam}.
Each GRU model in our study is reused in a UniTE-GEN model as the function-fitting component.

Based on preliminary experiments, we train each GRU and UniTE-DIS model for $10$ epochs with a batch size of $128$ trajectories.
The training trajectories are divided uniformly at random into $1163$ batches and are reused across all epochs.
The batches are shuffled before each epoch s.t.\ the are in random order.
For both UniTE-DIS and GRU, we selected the learning rate $\lambda$ by training a GRU model using for each learning rate $\lambda \in \{0.1, 0.01, 0.001, 0.0001\}$.
We use the learning rate $\lambda = 0.001$ with the best validation \gls{nll}.

For AGG, we selected aggregation threshold parameter $k$, contextual relevance parameter $c$, and the temporal relevance parameters $\delta$ using a grid search over values $k \in \{ 1, 2, 4, 8\}$, $c \in \{0, 1, 2, 4\}$ and $\delta \in \{15, 30, 60, 120 \}$.
We selected the hyperparameter configuration $k=1$, $c=0$, and $\delta=120$ with the best validation \gls{nll}.

The parameter $a$ used in the prior function layer (see \cref{alg:hybrid-output-layer}) serves a similar role as the aggregation threshold $k$ used in AGG.
We therefore set $a=1$ based on the best value $k=1$ for AGG.
We choose hyperparameters $c$ and $\delta$ using a grid search over the same values as for AGG.
For UniTE-DIS the best hyperparameters are $c=1$ and $\delta=120$, and $c=4$ and $\delta=15$ for UniTE-GEN.

\subsection{Performance Evaluation}\label{sec:results}
\todo{Update GRU performance increases}
We repeat our experiments ten times for each algorithm and report their mean performance with standard deviations in \cref{tab:results-overview}.
However, one of the GRU models did not finish properly. The numbers for GRU and UniTE-GEN in \cref{tab:results-overview} are therefore based on nine runs.
Since the AGG baseline is deterministic, no standard deviations are associated with its performance figures.

\begin{table}
  \caption{Algorithm performance on the test trajectories.\label{tab:results-overview}}
  \centering
  \begin{tabular}{llll}
    \toprule
    \emph{Algorithm} & \emph{\gls{nll}} & \emph{\gls{mae} (s)} & \emph{\gls{mape} ($\%$)} \\
    \midrule
    UniTE-DIS & $\mathbf{26.83} \boldsymbol\pm \mathbf{0.52}$ & $\mathbf{75.13} \boldsymbol\pm \mathbf{0.70}$ & $\mathbf{17.48} \boldsymbol\pm \mathbf{0.15}$ \\
    UniTE-GEN & $42.99 \pm 7.84$ & $83.35 \pm 1.66$ & $20.86 \pm 0.42$ \\
    GRU & $44.47 \pm 1.00$ & $97.38 \pm 2.77$ & $24.82 \pm 0.81$ \\
    AGG & $75.51 \pm 0.00$ & $77.09 \pm 0.00$ & $18.53 \pm 0.00$ \\
    \bottomrule
  \end{tabular}
  \todo{Update GRU and UniTE-GEN when final run finishes.}
\end{table}

\paragraph{Distribution Modeling}
As shown in \cref{tab:results-overview}, UniTE-DIS is the best-performing algorithm on average for both travel speed distribution modeling and travel time point estimation.
\todo{Update GRU performance difference}
On average, UniTE-DIS outperforms the GRU and AGG baselines by $39.68\%$ and $64.48\%$, respectively, on distribution modeling performance in terms of \gls{nll}.
UniTE-GEN also outperforms the baselines on distribution modeling, although not as substantially as UniTE-DIS, with improvements of $43.07\%$ and $3.33\%$ over AGG and GRU, respectively.
In addition, the results vary substantially between runs with a standard deviation of $7.48$.
Given these variations in \gls{nll} in combination with the low number of runs of UniTE-GEN, it is unclear whether UniTE-GEN outperforms GRU in general.

A detailed analysis of the results suggests that UniTE-GEN implicitly sacrifices estimation accuracy of the distribution mean for estimation accuracy of the distribution variance, a pattern not found in UniTE-DIS or the GRU baseline.
Specifically, when comparing any two runs of UniTE-DIS and the GRU baseline, the run with the best estimation of the distribution typically also has the distribution variance estimate.
In addition, the GRU model that provides the prior hyperparameters has been pre-trained on the data and therefore also to estimate travel speed distribution.
However, as a consequence of the rules for computing the posterior hyperparameters in \cref{eq:posterior-parameters}, the degrees of freedom of the estimated $t$-distribution of the posterior predictive increases with the number of historical records used.
This results in a reduced spread of the distribution, particularly for road segments with relatively few historical records that do not capture the travel speed distribution.

\paragraph{Travel Time Point Estimation}
Interestingly, the ranking of the algorithms w.r.t.\ travel time point estimation is different from the ranking w.r.t.\ distribution modeling, as shown in \cref{tab:results-overview}.
The differences between the algorithms are also smaller, with the GRU baseline being a notable exception.
UniTE-DIS remains best with average performance increases of $2.54\%$, $9.86$, and $22.85\%$ over AGG, UniTE-GEN, and GRU, respectively, in terms of \gls{mae}.
Unlike for the task of distribution modeling, UniTE-GEN provides a substantial performance improvement of $14.40\%$ over the GRU baseline for travel time point estimation, with a smaller standard deviation, i.e., $1.66$ vs. $2.77$ for \gls{mae}.

\paragraph{Summary}
UniTE-DIS performs {$39.68$--$64.48\%$} and {$2.54$--$22.85\%$} better on distribution modeling and travel time point estimation, respectively, than the function-fitting and aggregation-based baselines.
In addition, UniTE-DIS outperformed UniTE-GEN by {$9.86$--$37.60\%$} across all measures and does not suffer from the mean-variance estimation accuracy trade-off we observed in the UniTE-GEN model, leading to large variances in distribution modeling performance.
These findings show that, while perhaps unconventional, training UniTE models to optimize a discriminative objective rather than a conventional generative objective, is worthwhile.

\subsection{The Generalizability-Accuracy Trade-Off}
One of our primary goals for \gls{aha} is that \gls{aha} models are flexible s.t.\ they can exploit the generalizability of the function-fitting component in data-sparse situations and can exploit the accuracy of aggregation-based methods in data-abundant situations.
To investigate this capability, we collect the sNLL of different road segments during evaluation of the test trajectories.
Then, we group them by the number of historical records available according to the ground-truth arrival times at the segments and compute the mean sNLL for each of these groups using the least restrictive record selection strategy (AGG's).
  The number of historical records used by each algorithm may differ from this number, depending on the restrictiveness of the record selection strategy and the accuracy of the expected arrival time at the road segments, but they are very strongly correlated.
The relationship between mean sNLL and the number of historical records available is shown in \cref{fig:robustness}.

\begin{figure}
  \centering
  \begin{center}
    \scalebox{0.85}{
    \hspace*{-0.525cm}
    \input{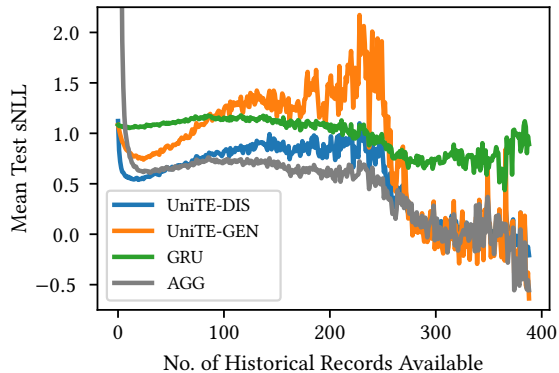}
  }
  \end{center}
  \vspace*{-0.35cm}
  \caption{The relationship between mean sNLL for road segment with different number of historical records available during evaluation on the test trajectories.
  \label{fig:robustness}}
\end{figure}

\COMMENT{
\begin{figure*}
  \centering
  \begin{subfigure}{0.495\textwidth}
    \hspace*{-0.525cm}
    \input{illustrations/robustness_zoomed_out.pgf}
    \caption{\label{fig:robustness-zoomed-out}}
  \end{subfigure}
  \begin{subfigure}{0.495\textwidth}
    \hspace*{-0.275cm}
    \input{illustrations/robustness_zoomed_in.pgf}
    \caption{\label{fig:robustness-zoomed-in}}
  \end{subfigure}
  \caption{Mean test sNLL for road segment with different numbers of historical records available.\label{fig:robustness}}
\end{figure*}
}

\paragraph{Analysis}
\cref{fig:robustness} shows that AGG has substantially worse performance than the other algorithms when few historical records are available, but that it overtakes GRU when $8$ historical records available.
Until about $35$ historical records are available, UniTE-DIS has the best performance, but from this point on, AGG has similar performance.

UniTE-DIS and AGG maintain quite similar performance when $80$ to $250$ historical records are available.
In this interval, the historical records have high quality, and there are sufficiently many.
The AGG baseline therefore overtakes UniTE-DIS and UniTE-GEN in terms of performance because it relies solely on the data and does not make use of a prior.
In addition, the record selection strategies of UniTE-DIS and UniTE-GEN are more restrictive, causing them to use, respectively, $34\%$ and $91\%$ less data on average than does AGG.
The adverse effect of using a prior in this interval is particularly strong for UniTE-GEN.\@
We expect this is (a) because it has the by far most restrictive record selection strategy and (b) because it expresses much higher certainty in the prior than UniTE-DIS does.\@
When more than $250$ historical records are available, UniTE-DIS and UniTE-GEN achieve similar performance to AGG since the influence of the prior becomes less significant.

\paragraph{Summary}
Both UniTE variants exhibit better generalizability than AGG when few historical records are available and achieve similar accuracy when many historical records are available, where
UniTE-DIS achieves superior generalizability and accuracy compared to UniTE-GEN.\@
Between these two extremes, AGG is superior to the UniTE variants due to the availability of sufficient high-quality data.
These data conditions are particularly favorable for AGG since it does not make use of a prior that may be inaccurate.

\subsection{Regularizing Properties}
An important property of the \gls{aha} framework is the implicit regularization w.r.t.\ data imbalances that results from the definition of the posterior.
In particular, the posterior (cf.\ \cref{eq:posterior-parameters}) implicitly regularizes the prior function since the influence of the prior function is inversely proportional to the number of historical records.
When training UniTE-DIS, we expect that this implicit regularization encourages the GRU architecture used as the prior function to output prior travel speed distributions that are accurate when only few historical records are available.
However, we do not expect this effect in GRU (or, equivalently, in UniTE-GEN) since historical records are not used at training time.




To study the regularization properties, we compare the segment-wise mean \gls{nll}, s\gls{nll}, of UniTE-DIS and GRU on the training set.
Let $e$ be a road segment that occurs $n$ times in the set of training trajectories.
Then, the sNLL of $e$ is
$\frac{1}{n}\sum_{i=1}^n - \ln p_i$,
where $p_i = \Pr(t_i \mid \mathbf{e}, \tau_i, \tilde{T}_i = \emptyset; \theta)$ is the value of the density function of the prior predictive at the $i$th occurrence.
Here, $\tau_i$ is the time of the occurrence, and $t_i$ is the ground truth travel speed.

\begin{figure}
  \begin{center}
    \scalebox{0.85}{
      \hspace*{-0.475cm}
      \input{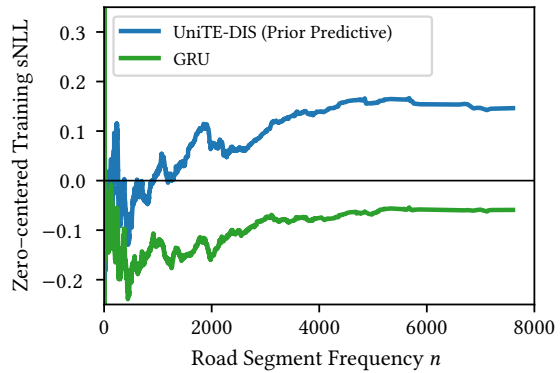}
    }
  \end{center}
  \vspace*{-0.35cm}
  \caption{Moving average (sample window size $\mathbf{250}$) of the zero-centered training sNLL of the prior predictive of UniTE-DIS and GRU at different road segment frequencies.\label{fig:regularization-analysis}}
\end{figure}

\cref{fig:regularization-analysis} plots the sNLL of all road segments as a function of their frequency.
For ease of comparison, we have centered the values around $0$ and use a moving average with a sample window size of $250$.
Thus, values above $0$ indicate below average performance and values below $0$ indicates above than average performance within the sample window.

As expected, the prior predictive of UniTE-DIS favors low-frequency road segments more than GRU and GRU favors high-frequency road segments more than UniTE-DIS.
Although not visible in \cref{fig:regularization-analysis}, the point of diversion occurs at a road segment frequency of $n=38$ corresponding to the $56$th percentile.
\christian{Hvis det er det vi skal se, hvorfor så ikke logaritmisk x-akse?}
This discrepancy continue to increase as the road segment frequency increases.
These findings suggest that the implicit regularization of the \gls{aha} framework contributes to the superior performance, shown in \cref{tab:results-overview}, of UniTE-DIS.

\subsection{Data Efficiency}
\begin{figure}
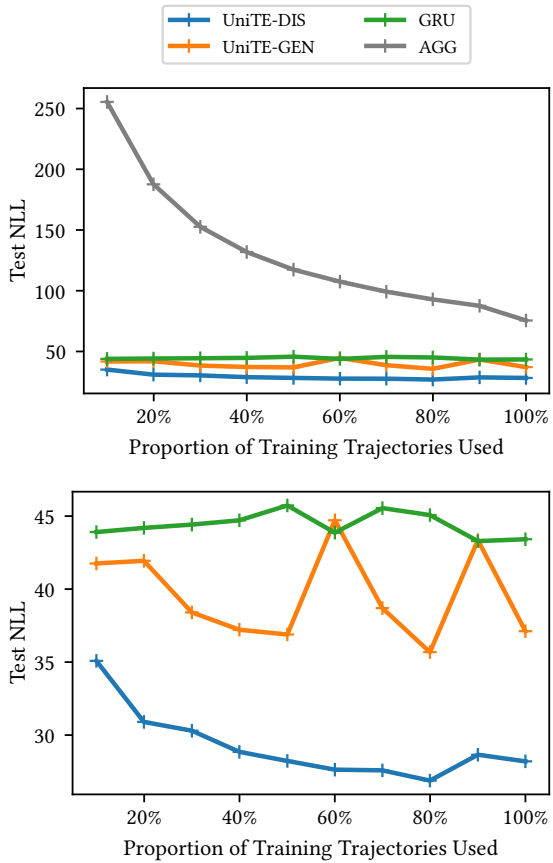

  \centering
    \vspace*{-0.65cm}
  \begin{subfigure}{\columnwidth}
  \begin{center}
    \scalebox{0.85}{
    \hspace*{-0.5cm}
    \input{illustrations/data_influence_nll_big.pgf}
  }
  \end{center}
  \end{subfigure}
  \begin{subfigure}{\columnwidth}
   \vspace*{-0.35cm}
  \begin{center}
    \scalebox{0.85}{
    \hspace*{-0.5cm}
    \input{illustrations/data_influence_nll_small.pgf}
  }
  \end{center}
  \end{subfigure}
   \vspace*{-0.35cm}
  \caption{Algorithm travel speed distribution modeling performance for different data subsets.\label{fig:data-influence}}
\end{figure}

We investigate how the performance of the algorithms changes depending on the training data available by giving them part of the training data, while keeping the number of iterations (i.e., backpropagations) constant.
For the sake of brevity, we show only the results in terms of test \gls{nll} in \cref{fig:data-influence}, but the patterns when using \gls{mae} and \gls{mape} are similar.

\paragraph{AGG and GRU}
As shown in \cref{fig:data-influence}, the performance of AGG is highly dependent on the size of the training set, whereas GRU is comparatively good at generalizing when using few training trajectories.
As discussed in \cref{sec:introduction-paper5}, function-fitting approaches such as GRU are good at generalization, and aggregation-based approaches such as AGG are not.
The results are therefore as expected.
However, it is notable that the performance of GRU is near-constant (within six standard deviations) and does not improve as more data becomes available.
We expect the primary cause to be the lack of high quality features available: using four attributes represented as $16$ features means that the feature space can be observed almost completely through few trajectories.

\paragraph{UniTE-GEN}
\cref{fig:data-influence} shows that UniTE-GEN tends to nearly-match or outperform it's pre-trained GRU component.
The results also suggest that UniTE-GEN tends to scale better with training data availability. However, the value of the UniTE framework when used in a generative manner is highly dependent on the pre-trained GRU model it uses.
As shown in the figure, this dependence results in a large performance variance that is consistent with the results in \cref{tab:results-overview}.
However, when measuring travel time point estimation using \gls{mae} or \gls{mape} (not shown), UniTE-GEN is strictly superior to its pre-trained function-fitting GRU component for all data set sizes and the performance difference increases proportionally to data set size.

\paragraph{UniTE-DIS}
UniTE-DIS outperforms the other algorithms for all data set sizes.
Unlike GRU, UniTE-DIS scales with data availability, although not as aggressively as AGG.
The results show that UniTE-DIS has high data efficiency and can substantially outperform a purely function-fitting and a purely aggregation-based approach even at very small data set sizes.
For instance, when using ca. $10\%$ of the training trajectories, UniTE-DIS outperforms AGG and GRU by $728\%$ and $20\%$, respectively.

\cref{fig:data-influence} suggests that the performance of UniTE-DIS deteriorates beyond $80\%$ of the training trajectories.
Given that the differences are small, we expect that this is due to the stochastic nature of the training process, but it may also be due to differences in the distributions of the training and test sets.
If the latter is the case, regularization techniques may be used during training to enhance generalizability.
However, since the performance differences are within six standard deviations, we cannot conclude which is the case.

\paragraph{Summary}
UniTE-DIS exhibits superior data efficiency compared to GRU and AGG and achieves superior performance for all data set sizes considered in our study and achieves superior performance at all data set sizes considered in our study.
And
Unlike the AGG baseline, UniTE-DIS exhibits good generalizability for small data set sizes.
And unlike the GRU baseline, the performance of UniTE-DIS improves proportionally to size of the data set.
UniTE-GEN exhibits the same behavior as UniTE-DIS when measuring travel time point estimation performance, albeit with strictly worse performance at all data set sizes.
However, when measuring distribution modeling performance, the potential performance increase of using UniTE-GEN on a pre-trained GRU model is highly dependent on the particular GRU model.

\subsection{Record Selection Strategies}
The value of the \gls{aha} framework depends on the record selection strategy.
In particular, there is a trade-off between data availability and data quality.
If the historical records are irrelevant, the posterior predictive may perform worse than the prior predictive.
However, if no historical records are available, \gls{aha} offers no benefits (but also no drawbacks).
In addition, we hypothesize that optimal record selection strategies for purely aggregation-based approaches differs from optimal selection strategies for \gls{aha}.
We therefore investigate how different values of the temporal relevance hyperparameter $\delta$ and the contextual relevance hyperparameter $c$ influence AGG, as well as UniTE-DIS and UniTe-GEN.
Recall that large $c$ values indicates high relevance and high $\delta$ indicates low relevance.

\begin{figure*}[ht]
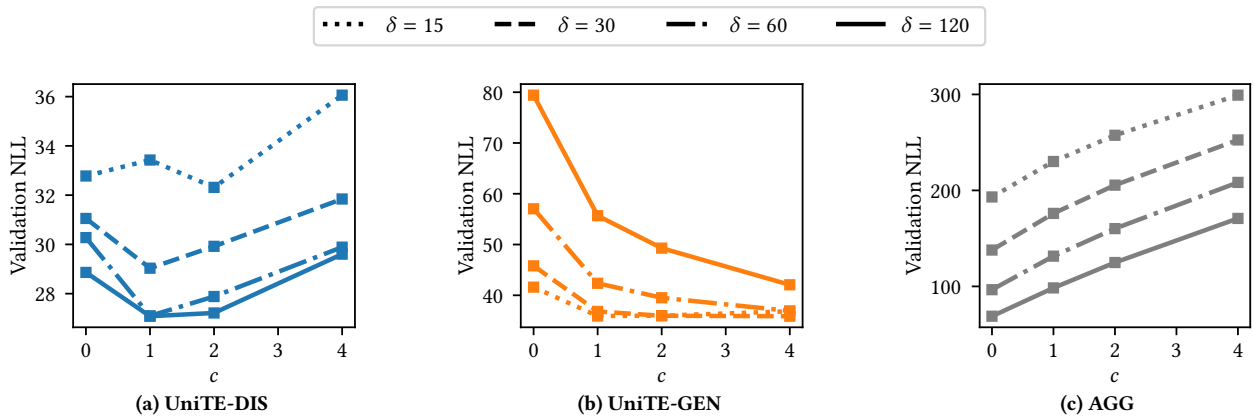

  \centering
  \begin{subfigure}{\textwidth}
    \centering
    \scalebox{0.85}{
      \input{illustrations/record_selection_legend.pgf}}
  \vspace*{-0.1cm}
  \end{subfigure}
  \begin{subfigure}{0.33\textwidth}
  \begin{center}
    \hspace*{-0.55cm}
    \scalebox{0.85}{
      \input{illustrations/record_selection_dis.pgf}}
  \vspace*{-0.55cm}
    \caption{UniTE-DIS}
  \end{center}
  \end{subfigure}
  \begin{subfigure}{0.33\textwidth}
  \begin{center}
    \hspace*{-0.55cm}
    \scalebox{0.85}{
      \input{illustrations/record_selection_gen.pgf}}
  \vspace*{-0.55cm}
   \caption{UniTE-GEN}
  \end{center}
  \end{subfigure}
  \begin{subfigure}{0.33\textwidth}
  \begin{center}
    \hspace*{-0.55cm}
    \scalebox{0.85}{
      \input{illustrations/record_selection_agg.pgf}}
  \vspace*{-0.55cm}
    \caption{AGG}
  \end{center}
  \end{subfigure}
  \caption{The distribution modeling performance on the validation set of (a) UniTE-DIS, (b) UniTE-GEN, and (c) AGG for different values of parameters $c$ and $\delta$ that regulate contextual and temporal relevance of the retrieved historical records.\label{fig:record-selection}}
\end{figure*}

\paragraph{Analysis}
\cref{fig:record-selection} shows the validation \gls{nll} found during hyperparameter selection to select the best combination of $c$ and $\delta$ values for AGG, UniTE-DIS, and UniTE-GEN.\@
As shown in the figure, the optimal values differ across the algorithms. 
In addition, they follow quite different patterns.

AGG benefits from high data availability even at the expense of data relevance.
This is not surprising given the inaccurate heuristic used to estimate the travel speed distributions when too few records are available.
UniTE-GEN follows the opposite pattern and prefers high data relevance.
This is likely caused by the variance-reducing effects of UniTE-GEN discussed in \cref{sec:results}, leading to too narrow travel speed distributions.
If few records are available, this effect is not as pronounced.
Finally, UniTE-DIS is somewhere in-between, preferring a moderate contextual relevance with $c=1$.
Temporal relevance is less important, with $\delta=60$ and $\delta=120$ yielding nearly equal performance, particularly when $c=1$.

\paragraph{Summary}
We find that optimal the record selection strategies differ across the \gls{aha} and AGG.
In particular, our results suggest that optimal strategies for \gls{aha} are more selective than those for aggregation-based approaches.
We attribute this to the difference in the quality of the `default' mechanism used to find travel speed distributions when few or no historical records are available.

\subsection{Processing and Storage Complexity}
Computational efficiency has not been the focus of our implementation efforts for UniTE or the baselines AGG and GRU.
However, we find it prudent to discuss the empirical efficiency of our approach as well as the theoretical complexity of \gls{aha}

\subsubsection{Empirical Performance}
Our implementation can train a UniTE-DIS model with approximately two seconds of processing time per trajectory using less than $15$ GB of RAM on a non-dedicated server with $2.3$ GHz processors.
For comparison, GRU takes approximately one and a half seconds of processing time per trajectory to train.
Once trained, both UniTE-DIS and UniTE-GEN can estimate the road segment travel speed distributions along a single trajectory in approximately $0.15$ seconds of processing time per trajectory.

\subsubsection{Complexity Analysis}
Estimation of road segment travel speed distributions along a single trajectory $\mathit{TR} = (\mathit{tr}_1, \dots, \mathit{tr}_n)$ through \gls{aha} requires $n$ historical record look-ups and $n$ computations of the posterior hyperparameters.

In the continuous time case, which we consider in our empirical study, look-ups can be performed efficiently by preprocessing the $N$ segment traversals across all trajectories.
The traversals may be stored in a list or an array, requiring $O(N)$ space, and can be sorted in $O(N\log N)$ time.
This enables the use of binary search to perform a look-up, requiring $O(\log N)$ time, but the search can be accelerated considerably in practice using indexing.
For instance, in our implementation, we use indexes to find historical records of a particular road segment from a particular day of the week in constant time and perform binary search only on this subset of historical records.
In general, each posterior hyperparameter computation can be done in $O(g(m))$ time, where $m$ is the number of historical records.
The function $g$ depends on the particular instance of \gls{aha}, and in the case of \gls{agha}, computation of the posterior parameters takes $O(m)$ time.
After preprocessing, the whole estimation procedure thus takes $O(n(\log N + g(m)))$ time in general and $O(n(\log N + m))$ for \gls{agha} with storage complexity $O(N)$ for a trajectory of length $n$.

Discrete time is commonly used in the literature, where time is partitioned into a set of fixed-size intervals $I$.
In this case, the estimation of the travel time or speed distributions along a trajectory can be reduced to $O(n|E||I|)$ time by computing and caching the posterior hyperparameters for each road segment and interval pair in $E \times I$ since, typically, $|E||I| << \log N + g(m)$.

\section{Related Work}\label{sec:related-work-paper5}
As discussed in \cref{sec:introduction-paper5}, approaches for travel time and speed estimation can be categorised broadly as either function-fitting%
~\citep{wei2020spatial,lu2020st,ge2020global,zhang2020novel,zhang2020graph,zhang2020network, yin2020attention,lee2020predicting,guo2019attention,cui2019traffic,yu2017spatio,htte,stad,zheng2013time,fu2020estimation,barnes2020bustr,lan2019travel,wu2019deepeta,shen2019tcl,hu2019stochastic,hu2020stochastic,taoyang2019deepist,xi2019path,compacteta,rade2018wedge,zheng2018learning,wang2018will,wang2014travel,yang2013using,workshop}
or aggregation-based%
~\citep{pace,hu2017enabling,dai2016path,yuan2011tdrive}
approaches.

Function-fitting approaches differ primarily in how they structure the function they are fitting, and aggregation-based approaches primarily differ in how they select or construct records for aggregation.
The details of these approaches are orthogonal to the novelty of the \gls{aha} framework.
The \gls{aha} framework can be used in conjunction with existing function-fitting as well as aggregation-based approaches, where existing function-fitting approaches can be used as the function-fitting component in \gls{aha} and existing aggregation-based methods can be used for record selection and construction in \gls{aha}.
To the best of our knowledge, \gls{aha} is the first approach that combines function-fitting and aggregation-based approaches in a single cohesive framework.
We expect that only minor modifications to the output and objective of a function-fitting approach is necessary, and that no changes to the record selection strategies of aggregation-based approaches are necessary.


Approaches to travel time and speed estimation can further be categorised as segment-based~\citep{htte,yang2013using,zheng2013time,barnes2020bustr,hu2019stochastic,rade2018wedge,hu2017enabling,wang2014travel,lee2020predicting,workshop}, route-based~\citep{fu2020estimation,pace,wu2019deepeta,shen2019tcl,lan2019travel,taoyang2019deepist,xi2019path,compacteta,zheng2018learning,dai2016path,wang2018will}, origin-destination based ~\citep{stad,hu2020stochastic,yuan2011tdrive}, or station-based~\citep{wei2020spatial,lu2020st,ge2020global,zhang2020novel,zhang2020graph,yin2020attention,zhang2020network,guo2019attention,cui2019traffic,yu2017spatio}, meaning that they target estimation for segments, routes, origin-destination pairs, and traffic measuring stations, respectively.
Traffic measuring stations typically represent loop detectors in traffic forecasting applications.

In this paper, we have adopted a segment-based and a route-based perspective, but the \gls{aha} framework only requires that the type of data instance, be it a segment, route, origin-destination pair, or a measuring station, is represented as a feature vector.
The framework is thus equally compatible with origin-destination-based and station-based approaches, given that appropriate methods of feature vector construction and record selection are available.

\section{Conclusions and Future Work}\label{sec:epilogue-paper5}
We have presented \gls{aha}, a novel framework that provides a Unifying Approach to Travel time and speed Estimation. \gls{aha} unifies function-fitting and aggregation-based approaches to travel time and speed estimation to leverage the generalizability of function-fitting approaches with the accuracy of aggregation-based approaches.
By virtue of being a Bayesian framework, \gls{aha} is able to switch smoothly between its constituent function-fitting and aggregation-based components depending on data availability.

In our empirical study, we found that \gls{aha} can improve the accuracies of travel speed distribution and travel time estimation by {$40$--$64\%$} and {$3$--$23\%$}, respectively, compared to using function fitting or aggregation alone.
These improvements result from the superior generalizability relative to both the function-fitting approach and the  aggregation-based approach in our study, while maintaining superior or similar accuracy relative to the aggregation-based approach across all data availability scenarios in our dataset.

\gls{aha} has a number of other benefits in addition to estimation performance improvements.
First, the framework implicitly regularizes its function-fitting component to handle issues of data imbalance and reduce model bias due to its Bayesian nature.
Second, \gls{aha} models are less reliant on the structural quality of both the mapping function, i.e., the neural network architecture in our empirical study, and input feature vector representations since they also use aggregation during estimation.
This property of \gls{aha} can reduce the typically substantial resources required for feature engineering and neural network architecture design when using neural networks for function-fitting.

Future directions for \gls{aha} include exploring more complex models of travel time and speed distributions.
This may necessitate more sophisticated techniques than the conjugate Bayesian analysis~\citep{murphy2007conjugate} used in \gls{agha}, e.g., variational inference techniques~\citep{blei2017variational}.
In addition, investigating further synergistic effects of function-fitting and aggregation within the \gls{aha} framework is of interest.
For instance, in our empirical study, the internal state of the recurrent \gls{gru} cell used in the \gls{aha} models is unaffected by the computation of the posterior.
This makes it more difficult for \gls{gru} cells to leverage correlations in travel speed between adjacent road segments in a route for better estimation accuracy.

\section*{Appendices}
\begin{appendices}
  \crefalias{section}{appendix}
  \section{Definition of AGG}\label{app:aggregation-baseline-paper5}
The distribution derivation process when no historical records are available is not clear from the literature, but Hu et al.~\citep{hu2017enabling} suggests that they use the speed limit as a deterministic (rather than probabilistic) travel speed estimate.
A direct application of this approach to our setting results in a Gaussian with the speed limit as the mean and (near-)zero variance.
However, such a low variance is unrealistic and severely decreases the travel speed distribution modeling performance of AGG.
To achieve a fair comparison, we therefore do the following.

Given a road segment $e_i$, AGG outputs the mean $\mu_i$ and the standard deviation $\sigma_i$:
\begin{equation}\label{eq:aggregation-baseline-outputs}
  \begin{split}
    \mu_i &= \begin{cases}
      M_{\tilde{T}_i} & \text{if } |\tilde{T}_i| \geq k \\
       0.79\cdot\text{SL}(e_i) & \text{otherwise} \\
      \end{cases} \\
      \sigma_i &= \begin{cases}
        S_{\tilde{T}_i} & \text{if } |\tilde{T}_i| \geq k \text{ and } |\tilde{T}_i| > 1 \\
        0.07\cdot\mu_i & \text{otherwise}
      \end{cases}
  \end{split}
\end{equation}
Here, $M_{\tilde{T}_i}$ and $S_{\tilde{T}_i}$ are the arithmetic mean and the standard deviation of historical records $\tilde{T}_i$, respectively.
The function $\text{SL}$ returns the speed limit for its argument road segment $e_i$.

The factors $0.79$ and $0.07$ used in \cref{eq:aggregation-baseline-outputs} to derive a mean and standard deviation when the number of historical records is insufficient are chosen based on our knowledge of the domain and the dataset, and takes into account that vehicles tend to travel at speeds below the speed limit on urban roads~\citep{yang2013using}, which occur occur in the data set used in our study.
As an example, if the speed limit is $50$ km/h then drivers drive at around $40$ km/h on average, and $99.7\%$ of drivers are expected to drive below the speed limit (as a consequence of the empirical rule).
From our experience, this scenario is quite realistic, and using of $79\%$ (rather than $100\%$) of the speed limit as the distribution mean yielded substantial performance improvements in terms of travel time point estimation in preliminary experiments.

\subsection{Speed Limit Derivation}
The function invocation $\text{SL}(e_i)$ in \cref{eq:aggregation-baseline-outputs} returns the speed limit of road segment $e_i$ if it exists in our dataset.
However, as noted in \cref{sec:data-set-paper5}, the dataset used in our study does not contain a speed limit for all road segments.
When no speed limit is given, $\text{SL}(e_i)$ instead returns a speed limit derived from road segment $e_i$'s attributes using a \gls{osm} heuristic\footnote{\url{https://wiki.openstreetmap.org/wiki/OSM_tags_for_routing/Maxspeed\#Default_speed_limits}}.

Since our data is from Denmark, we use the \gls{osm} speed limit heuristic for Denmark. It is as follows.
\begin{enumerate}
    \item If the road category of a road segment is motorway then assign a speed limit of $130$.
    \item If the road category is trunk then assign a speed limit $80$.
    \item If the road category is neither motorway or trunk, but the road segment is within a city, then assign a speed limit of $50$.
    \item Otherwise, assign a speed limit of $80$.
\end{enumerate}
A road segment is considered to be in a city if either the source intersection or the target intersection of the road segments is in a city.

\subsection{Record Selection}\label{app:record-selection}
The AGG baseline relies on a record selection strategy to find historical records $\tilde{T}_i$ to compute the sample mean and sample standard deviation used in \cref{eq:aggregation-baseline-outputs}.
The algorithm used for record selection is presented in \cref{alg:record-selection}.

The algorithm takes as input a set of training trajectories $\mathcal{TR}$, the route $p$ for which the travel time is in the process of being estimated, the road segment $e_i \in p$ for which historical records $\tilde{T}_i$ are currently being collected, and the arrival time $\tau_i$ at road segment $e_i$.
In addition, the algorithm takes as input two parameters: an integer contextual relevance parameter, $c$, and a temporal relevance parameter, $\delta$, in some unit of time.
Higher values of $c$ returns fewer, but more relevant historical records.
Higher values of $\delta$ returns more, but less relevant historical records.

\begin{algorithm}
  \caption{Record Selection Algorithm\label{alg:record-selection}}
  \begin{algorithmic}[1]\raggedright
    \Function{RecordSelection}{$\mathcal{TR}$, $p=(e_1, \dots, e_q)$; $e_i$, $\tau_i$, $c$, $\delta$}
        \State{$\tilde{T}_i \gets \emptyset$}
        \State{$C_i \gets (e_{i-c}, \dots, e_{i-1}, e_i, e_{i+1}, \dots, e_{i+c})$}\label{line:record-selection-context-i}
          \State{$I_i \gets [\tau_i - \frac{\delta}{2}; \tau_i + \frac{\delta}{2}]$}
        \ForEach{trajectory $\mathit{TR} \in \mathcal{TR}$}
            \State{\Let{$\mathit{TR} = \big((e'_1, \tau_1, t_1), \dots, (e'_n, \tau_n, t_n)\big)$}}
            \For{$j = 1$ to $n$}
                \State{$C_j \gets (e'_{j-c}, \dots, e'_{j-1}, e'_j, e'_{j+1}, \dots, e'_{j+c})$}
                \If{$C_i = C_j \land \tau_j \in I_i \land t_j \neq \varnothing$}
                    \State{$\tilde{T}_i \gets \tilde{T}_i \cup \{t_j  \}$}
                \EndIf
            \EndFor
        \EndFor
        \State{\Return{$\tilde{T}_i$}}
    \EndFunction
  \end{algorithmic}
\end{algorithm}

\cref{alg:record-selection} constructs the set of historical records $\tilde{T}_i$ as follows.
The algorithm scans the set of trajectories in the loop in {Lines~$5$--$10$} for historical records.
For each trajectory, the algorithm scans each traversal in the trajectory for historical records in the loop in {Lines~$7$--$10$}.
A historical record refers strictly to the recorded travel time or travel speed of the traversal.

To be selected, a historical record must satisfy the three conditions in Line~9.
First, it must be \emph{contextually relevant} s.t.\ the contexts of road segments $e_i$ and $e'_j$ are identical, i.e., $C_i = C_j$.
Here, context refers to the preceding and succeeding road segments of a road segment in a trajectory or a route.
Second, the historical record must be \emph{temporally relevant}, i.e., occur at a similar time of week as $\tau_i$, defined by the interval $I_i$ (Line~4).
Finally, the historical record is added to $\tilde{T}_i$ if $t_j \neq \varnothing$, i.e., if the historical record is derived from \gls{gps} data and is not created by the map-matching algorithm.

\section{Definition of GRU}\label{app:gru}
We express GRU in the \gls{aha} framework as the prior function $f$.

Let $p=(e_1, \dots, e_n)$ be the input route starting at time $\tau_1$.
For each road segment $e_i$ in $p$, \gls{gru} computes the following.
\begin{equation}\label{eq:gru}
  \begin{split}
    \mathbf{x}_i &= \mathbf{e}_i \oplus \boldsymbol\tau_i \\
    \mathbf{z}_i &= \text{GRU}(\mathbf{x}_t, \mathbf{z}_{i-1}) \\
    \mathbf{h}_i &= \mathbf{z}_i \oplus \mathbf{x}_i \\
    f(\mathbf{e}_i, \boldsymbol\tau_i; \psi) &= \text{PriorFunctionLayer}(\mathbf{h}_i),
  \end{split}
\end{equation}
where $z_0$ is a $d$-dimensional vector of zeros.
For $i > 1$, we compute $\tau_i$ by incrementing $\tau_{i-1}$ by the expected time to traverse road segment $e_{i-1}$, i.e., we increment $\tau_{i-1}$ by $\frac{l_{i-1}}{\mu_{i-1, m}}$ where $l_{i-1}$ is the length of road segment $e_i$ and $\mu_{i-1, m}$ is the expected travel speed when traversing road segment $e_i$ (calculated using \cref{eq:posterior-parameters}).
The prior function layer is described in \cref{alg:hybrid-output-layer}.
During training, we use the time $\tau_i$ recorded during the input training trajectory if $\tau_i \neq \varnothing$.

As shown in \cref{eq:gru}, \gls{gru} takes as input the $32$-dimensional vector $\mathbf{x}_i$, a concatenation of the $16$-dimensional vector representations of road segment $e_i$ and $\tau_i$.
We explain how $\boldsymbol\tau_i$ is constructed in \cref{app:representation-of-time}.
Vector $\mathbf{x}_i$ is passed to a \gls{gru} cell that outputs a $32$-dimensional vector $\mathbf{z}_i$.
The \gls{gru} cell is recurrent and therefore takes as input vector $\mathbf{z}_{i-1}$, the output of the \gls{gru} cell at the previous road segment of the route.

Preliminary experiments indicated that the use of a \emph{skip connection} is beneficial to the \gls{gru} baseline, i.e., a connection from an earlier layer to later layer with at least one layer in-between.
The skip connection is captured in the computation of $\mathbf{h}_i$ in \cref{eq:gru} where the output of the \gls{gru} cell $\mathbf{z}_i$ is concatenated with the input vector $\mathbf{x}_i$.
Finally, \gls{gru} applies the prior function layer described in \cref{alg:hybrid-output-layer} to vector $\mathbf{h}_i$ to output the prior hyperparameters.
Note that $\tilde{T}_i = \emptyset$ when computing the posterior predictive $\Pr(\hat{t}_i=t_i \mid \tilde{T}_i, \theta_i)$ of the \gls{gru} baseline, e.g., when computing \gls{nll} (cf.\ \cref{eq:nll}) during training or evaluation.

\subsection{Representation of Time}\label{app:representation-of-time}
The representation of time $\boldsymbol\tau_i$ used by the \gls{gru} algorithm in \cref{eq:gru} is a $16$-dimensional feature vector representation of the time of week, where $8$ dimensions are used to represent the time of day and $8$ dimensions are used to represent the day of the week.
Formally, we learn a time-of-week vector
$
 \boldsymbol\tau = \boldsymbol\tau_{\mathit{tod}} \oplus \boldsymbol\tau_{\mathit{dow}}
$
for time $\tau$, where $\boldsymbol\tau_{\mathit{tod}} \in \mathbb{R}^8$ represents the time of day, $\boldsymbol\tau_{\mathit{dow}} \in \mathbb{R}^8$ represents the day of the week, and $\oplus$ denotes vector concatenation.
Preliminary experiments indicated that our results are robust to changes to the dimensionality of this vector representation.

To represent time of day in our experiments, we divide the time of day into $96$ $15$-minute intervals $\mathcal{I} = \{I_1, \dots, I_{96} \}$ s.t.\ $I_1 = \interval{0:00}{0:15}$, $I_2 = \interval{0:15}{0:30}$, and so forth.
Then, we one-hot encode the time of day $\tau_{\mathit{tod}}$ into a $96$-dimensional vector $\boldsymbol\tau'_{tod} = \begin{bmatrix} \mathbbm{1}[\tau_{\mathit{tod}} \in I_1] & \dots & \mathbbm{1}[\tau_{\mathit{tod}} \in I_{96}] \end{bmatrix}$, where $\mathbbm{1}$ is the indicator function.
Finally, we multiply the one-hot encoding $\boldsymbol\tau'_{\mathit{tod}}$ by a trainable matrix $\mathbf{W}_{\mathit{tod}} \in \mathbb{R}^{96 \times 8}$ to achieve the time of day representation
  $\boldsymbol\tau_{\mathit{tod}} = \boldsymbol\tau'_{\mathit{tod}}\mathbf{W}_{\mathit{tod}}$ with dimensionality $8$.

We represent the day of week in a manner similar to the time of day, but with $7$ dimensions in the one-hot encoding $\boldsymbol\tau'_{\mathit{dow}}$---one for each day of the week---and multiply $\boldsymbol\tau'_{\mathit{dow}}$ by a trainable weight matrix $\mathbf{W}_{\mathit{dow}} \in \mathbb{R}^{7 \times 8}$ to get an $8$-dimensional vector representation $\boldsymbol\tau_{\mathit{dow}}$ of the day of the week.

\end{appendices}
\clearpage
\printbibliography
\end{document}